\documentclass[preprint,12pt]{elsarticle_nonatbib}

\usepackage{amssymb}
\usepackage{lineno}

\usepackage{caption}
\usepackage{subcaption}

\usepackage[colorlinks=true,linkcolor=blue,urlcolor=black,bookmarksopen=true]{hyperref}
\usepackage{bookmark}

\usepackage{adjustbox}
\usepackage{float}

\usepackage[T1]{fontenc}
\usepackage{libertine}
\usepackage{amsmath}
\usepackage[libertine]{newtxmath}
\usepackage{tikz}
\usetikzlibrary{arrows,backgrounds,positioning,shapes,fit}
\usepgflibrary{shapes.multipart}
\usepackage{algpseudocode}
\usepackage{algorithm,tabularx}
\usepackage[T1]{fontenc}
\usepackage{pgfplots}
\usepackage{multirow}
\usepackage{siunitx}
\usepackage{threeparttable}     % table notes
\usepackage{makecell}
\usepackage{booktabs}
\usepackage{caption}
\usepackage{siunitx}
\usepackage{xcolor}

\newcolumntype{L}[1]{>{\raggedright\arraybackslash\hspace{0pt}}m{#1}}
\newcolumntype{C}[1]{>{\centering\arraybackslash\hspace{0pt}}m{#1}}

\definecolor{lightgray}{rgb}{0.9, 0.9, 0.9}

\usetikzlibrary{shapes.geometric,matrix,chains,positioning,decorations.pathreplacing,arrows,circuits.ee.IEC}

\def\HiLi{\leavevmode\rlap{\hbox to \hsize{\color{yellow!50}\leaders\hrule height .8\baselineskip depth .5ex\hfill}}}

\pgfplotsset{
	compat=newest,
	xlabel near ticks,
	ylabel near ticks
}

\tikzstyle{startstop} = [rectangle, rounded corners, minimum width=3cm, minimum height=1cm,text centered, draw=black, fill=red!30]
\tikzstyle{io} = [trapezium, trapezium left angle=70, trapezium right angle=110, minimum width=3cm, minimum height=1cm, text centered, text width=3cm, draw=black, fill=blue!30]
\tikzstyle{process} = [rectangle, minimum width=3cm, minimum height=1cm, text centered, draw=black, fill=orange!30]
\tikzstyle{decision} = [diamond, minimum width=3cm, minimum height=1cm, text centered, draw=black, fill=green!30]
\tikzstyle{arrow} = [thick,->,>=stealth]

\algnewcommand\algorithmicforeach{\textbf{for each}}
\algdef{S}[FOR]{ForEach}[1]{\algorithmicforeach\ #1\ \algorithmicdo}

\makeatletter
\newcommand{\multiline}[1]{%
  \begin{tabularx}{\dimexpr\linewidth-\ALG@thistlm}[t]{@{}X@{}}
    #1
  \end{tabularx}
}
\makeatother

\journal{TBD}

\usepackage[
backend=biber,
style=apa,
]{biblatex}

\begin{document}

\def\btc{\begin{tabular}{c}}
\def\etc{\end{tabular}}

\begin{frontmatter}

\title{A Neuromorphic Architecture for Reinforcement Learning from Real-Valued Observations}

% \tnotetext[n1]{Funding: This work was supported by the FACEPE, CNPq  and CAPES (Brazilian Research Agencies).}

\auth[add1]{S\'ergio F. Chevtchenko\corref{cor1}}
\ead{sfc@cin.ufpe.br}
\cortext[cor1]{Corresponding author}
\auth[add2]{Yeshwanth Bethi}
\ead{y.bethi@westernsydney.edu.au}
\auth[add1]{Teresa B. Ludermir}
\ead{tbl@cin.ufpe.br}
\auth[add2]{Saeed Afshar}
\ead{S.Afshar@westernsydney.edu.au}
\address[add1]{Centro de Inform\'atica - CIn, Universidade Federal de Pernambuco, Av. Jornalista An\'ibal Fernandes, s/n, Cidade Universit\'aria, 50.740-560, Brazil}
\address[add2]{International Centre for Neuromorphic Engineering, MARCS Institute, Western Sydney University,
Werrington, NSW 2747, Australia}

\begin{abstract}
Reinforcement Learning (RL) provides a powerful framework for decision-making in complex environments. However, implementing RL in hardware-efficient and bio-inspired ways remains a challenge. This paper presents a novel Spiking Neural Network (SNN) architecture for solving RL problems with real-valued observations. The proposed model incorporates multi-layered event-based clustering, with the addition of Temporal Difference (TD)-error modulation and eligibility traces, building upon prior work. An ablation study confirms the significant impact of these components on the proposed model's performance. A tabular actor-critic algorithm with eligibility traces and a state-of-the-art Proximal Policy Optimization (PPO) algorithm are used as benchmarks. Our network consistently outperforms the tabular approach and successfully discovers stable control policies on classic RL environments: mountain car, cart-pole, and acrobot. The proposed model offers an appealing trade-off in terms of computational and hardware implementation requirements. The model does not require an external memory buffer nor a global error gradient computation, and synaptic updates occur online, driven by local learning rules and a broadcasted TD-error signal. Thus, this work contributes to the development of more hardware-efficient  RL solutions.
\end{abstract}


\end{frontmatter}

%\newpage
% \linenumbers

\section{Introduction}
Reinforcement Learning (RL) is a bio-inspired paradigm within machine learning, where an agent learns to make decisions by interacting with an environment and receiving feedback in the form of rewards \parencite{sutton2018reinforcement}. It has been successfully used in various domains, ranging from game-playing AI \parencite{mnih2015human}  to self-driving cars, robotics \parencite{hwangbo2019learning}, and much more. In recent years, deep reinforcement learning (DRL) has emerged as a promising approach, combining RL with the scalability and representation power of deep neural networks (DNNs). However, despite these advancements, the computational efficiency of deep neural networks still remains a challenge. Training DRL systems often require thousands of hours of training on energy-intensive computational resources like GPUs and TPUs.

An alternative approach is offered by biologically inspired, spiking or event-based, neural networks. These networks aim at emulating the computational principles found in biological brains, such as the propagation of information through discrete spikes, distributed spatio-temporal processing and local learning rules via co-located memory and computing blocks. Although bio-inspired methods have the potential to mitigate some of the challenges encountered by DNNs—particularly in the area of power efficiency and online learning—the use of SNNs for complex RL problems is still an area of active research\parencite{akl2023toward}.

In this paper, we introduce a novel neuromorphic architecture designed to tackle reinforcement learning problems with real-valued observations. Drawing inspiration from prior works \parencite{afshar2020event, BethiEtAl2022}, our network incorporates event-based clustering and introduces modulation by a global signal through eligibility traces. The effectiveness of the proposed model is assessed against a tabular actor-critic algorithm with eligibility traces and a state-of-the-art DRL model, Proximal Policy Optimization (PPO). The results demonstrate that our network provides an appealing trade-off in terms of computational and hardware implementation requirements when compared to PPO on three classic RL control tasks.

This paper is structured as follows. A review of related literature and comparison to the present work is provided in Section \ref{sec_related}. The proposed model is described in Section \ref{sec_proposed_SNN} and is experimentally compared to baseline models in Section \ref{sec_exp_eval}. Concluding remarks and discussion of possible future endeavors can be found in Section \ref{sec_conclusion}. %Additional experiments and code are provided as supplementary materials.

\section{Related works}
\label{sec_related}

Deep learning has shown impressive results in the domain of Reinforcement Learning and has surpassed all other conventional methods to be the state-of-the-art architectures for training autonomous agents, particularly on sparse reward structures. However, the training procedures for Artificial Neural Networks (ANNs) have always been resource-intensive in terms of computation and memory requirements. Deploying ANNs at the edge with low-power constraints and training them online still remains a challenge due to the same reason. Recent advances in neuromorphic computing and training Spiking Neural Network (SNN) architectures offer novel solutions to building low-power machine intelligence. SNN architectures that use binary valued spikes (or events) for both computation and communication provide excellent energy efficiency benefits. With the uptake in the adoption of neuromorphic sensors like DVS \parencite{LichtsteinerEtAl2008} and ATIS \parencite{PoschEtAl2010}, event-driven machine learning algorithms are also being explored for low-latency and low-power applications of neuromorphic computing \parencite{GallegoEtAl2020}. 

Training SNN architectures is still an active area of research, and many attempts have been made to approximate the error backpropagation and gradient descent techniques used to train ANNs and train SNNs. For example, \Textcite{bellec2020solution} proposed a spiking neural network learning rule called e-prop that can be applied to recurrent spiking networks. This rule is shown to approximate the performance of an LSTM network, trained via backpropagation through time on two discrete Atari games. Other methods use surrogate gradients with stand-in differentiable alternatives for non-differentiable parts of SNN architectures \parencite{NeftciEtAl2019}. For instance, \Textcite{akl2023toward} propose a combination of DRL frameworks with a spiking architecture through the use of surrogate gradients and the backpropagation through time (BPTT) algorithm. However, these methods also face the same problem of requiring energy-intensive computational resources to perform the error backpropagation. The error back-propagation approximations are also not bio-plausible as they would require symmetric backward pathways that transfer precise continuous-valued gradients and, in some cases, non-causal operations like Back-Propagation Through Time (BPTT). Bio-plausible local learning rules that can train individual nodes in SNN architectures using only information available at each neuron, can offer a solution to this problem. Spike-Timing-Dependent Plasticity (STDP) has been the most commonly used local learning rule for SNN architectures to perform unsupervised learning. STDP can train spiking neurons to learn the spatio-temporal features that reflect the statistics of the input spiking data. The Hebbian learning rule has primarily been used in unsupervised learning settings to learn some useful features for tasks like pattern classification \parencite{DiehlCook2015}.  However, unsupervised learning rules like STDP neglect any information related to the ``success'', ``failure'', or ``novelty'' of the inputs and outputs \parencite{2016_Fremaux_Neuromodulated_STDP}. 

R-STDP (Reward-modulated Spike Timing-Dependent Plasticity) is a Spiking Neural Network (SNN) learning rule that governs the changes in the strength of connections between neurons such that the plasticity of synapses is modulated by reward signals that reinforce or weaken the connections between neurons based on the timing of their spikes. \Textcite{2007_Izhikevich_Reward}, \Textcite{2007_Florian_Reinforcement} and \Textcite{legenstein2008learning} independently laid the groundwork for R-STDP modulated spiking networks. In order to demonstrate the computational and temporal capabilities of this approach, a fully connected multilayered network of spiking neurons is shown by \Textcite{2007_Florian_Reinforcement} to solve a temporally coded XOR problem with a delayed reward. \Textcite{VasilakiEtAl2009} used a reward-modulated STDP learning rule on the action layer in combination with an input place cell layer to perform reinforcement learning on a Morris water maze puzzle. While theoretical models have used reward-modulated STDP for over a decade, recent research has provided experimental evidence of the role of eligibility traces in Reinforcement Learning (RL)\parencite{2018_Gerstner_Survey_eligibility}. The R-STDP rule used in the present work is a simplified version of RMSTDPET found in the work by \Textcite{2007_Florian_Reinforcement}. Actor-critic models with temporal difference (TD) are proposed by \Textcite{2011_Potjans_Imperfect_TD}, and later  \Textcite{2013_Fremaux_RL_AC_SNN}. A more general three-factor learning rule is later introduced by \Textcite{2016_Fremaux_Neuromodulated_STDP}. 

A key component of reinforcement learning is the development of useful internal representations of complex environments to evaluate and utilize the current state of the environment. This allows the system to decide on actions that provide maximum future rewards \parencite{sutton2018reinforcement}. Various internal representations for inputs have been used by the SNN solutions for reinforcement learning. One strategy is to use manually selected partitions of the input space to generate spiking inputs for various states. \Textcite{2009_Potjans_Spiking_AC} and \Textcite{JitsevEtAl2012} used a fixed cluster of neurons that represent the state space and individually fire for each particular state. \Textcite{2013_Fremaux_RL_AC_SNN} used pre-determined place cells to act as input to their continuous time spiking actor-critic architecture. \Textcite{FriedrichEtAl2014} used population coding for the input layer representation. These methods manually partition the input space or use a population that uniformly covers the entire input space of any given environment. A drawback of these early models is that the implemented networks fully encode the observed state in the input layer, functionally similar to classical tabular RL algorithms. In other words, each neuron in the input layer is used to encode a specific state of the environment. Also, this approach does not scale well with increasing input dimension.  This limitation in scalability is addressed in our previous paper \parencite{chevtchenko2020learning} by a four-layered network with a hidden layer inspired by place cells. However, this approach often requires a large number of place neurons and relies on discrete and finite observations.

Other approaches to represent the input space include using the reservoir computing paradigm (\Textcite{SchrauwenEtAl2007};\Textcite{LukovsevivciusJaeger2009}). Reservoir computing involves projecting the low dimensional inputs to a high dimensional representational space by using a population of recurrently connected neurons such that states not linearly separable in the input space can be separated in the representational space \parencite{MaassEtAl2002}. \Textcite{WeidelEtAl2021} used reservoir computing with unsupervised plasticity on the inputs and apply reward modulated learning on the output layer that generated actions. However, most of these SNN architectures predominantly use rate coding, which does not offer the same level of energy efficiency that sparse temporal coding provides.

Another promising approach towards realizing low-power spike-based learning algorithms lies in the employment of event-driven neuromorphic algorithms. These are principally aimed at hardware deployment and aim for improved scalability through the collocation of processing and memory, as well as facilitating low-latency computation \parencite{SchumanEtAl2022}. Neuromorphic event-driven algorithms emulate spiking neuron architectures that are tailored for computational efficiency and performance. HFirst algorithm \parencite{OrchardEtAl2015} is an example of a multi-layered network architecture that is based on the HMAX  algorithm \parencite{SerreEtAl2007} which itself inspired by visual cortex to learn features from event-based data using an event-driven approach of processing data. The Hierarchy of Event-based Time Surfaces (HOTS) \parencite{LagorceEtAl2016} is another unsupervised multi-layered feature extraction algorithm that uses the same neuron update rule based on the cosine distance of weights to the input representation introduced in \Textcite{BallardEtAl2012} to model sensory features in cortical neurons. The HOTS algorithm uses multiple layers to extract spatio-temporal patterns at multiple hierarchical levels with different time constants by performing clustering. \Textcite{afshar2020event} introduced an unsupervised feature extraction algorithm called Feature Extraction using Adaptive Selection Thresholds (FEAST), which is also capable of learning hierarchical spatio-temporal features by using adaptive thresholds for each neuron to promote equal activation of neurons. 

The majority of the neuromorphic feature extraction algorithms are primarily targeted toward performing tasks on event-based data. The FEAST method has been used for a range of applications such as event-based object tracking \Textcite{RalphEtAl2022}, activity-driven adaptation in SNNs \Textcite{HaessigEtAl2020}, and spoken digits recognition task \Textcite{YingEtAl2022}. Recently a generalization of the FEAST algorithm has been proposed in \Textcite{BethiEtAl2022} to perform supervised learning on spiking data using reward and punishment of neurons through threshold adaptation. We can utilize the same neuromorphic principles and apply these architectures to the reinforcement learning domain to partition the input state space that can dynamically adapt to the reward structure and performance of the agent. In this work, we modify the neuron layers proposed in the \Textcite{BethiEtAl2022} to simultaneously perform unsupervised clustering of the input space and modulate it based on the ongoing TD-error of an actor-critic agent that utilizes the learned representation. We show that partitioning achieved by the neurons represents different parts of the input space with varying degrees of density and resolution depending on the ongoing performance of the actor-critic agent using the representation. We demonstrate the quality of the input space representations by using a conventional tabular actor-critic algorithm and applying it to various environments.

\section{The proposed architecture}
\label{sec_proposed_SNN}

The proposed network has two main components: i) an initial clustering layer(s) for dimensionality reduction and partitioning of the input space and ii) actor-critic neurons for learning from the previously reduced state-space representation. The following sections provide a detailed description of each layer and corresponding plasticity rules.

\subsection{Input layer}
\label{ssec_proposed_SNN_input}

\begin{figure}[H]
\centering
\begin{tikzpicture}[node distance=2cm]
    % Input signal
    \node (x1) {$x_1$};
    \node (x2) [below=0.5cm of x1] {$x_2$};
    \node (dots) [below=0.5cm of x2] {$\vdots$};
    \node (xn) [below=0.5cm of dots] {$x_n$};

    % Neurons
    \node[draw, circle, minimum size=0.8cm] (n1) [right=3cm of x1] {$n_1$};
    \node[draw, circle, minimum size=0.8cm] (n2) [right=3cm of x2] {$n_2$};
    \node (ndots) [below=0.3cm of n2] {$\vdots$};
    \node[draw, circle, minimum size=0.8cm] (ni) [right=3cm of xn] {$n_i$};

    % Threshold parameters
    \node (theta1) [right=0.1cm of n1] {$\theta_1, v_1$};
    \node (theta2) [right=0.1cm of n2] {$\theta_2, v_2$};
    \node (thetai) [right=0.1cm of ni] {$\theta_i, v_i$};

    % Synaptic weights
    \draw[->] (x1) -- (n1) node[midway,above] {$w_{11}$};
    \draw[->] (x1) -- (n2) node[midway,above] {$w_{12}$};
    \draw[->] (x1) -- (ni);
    \draw[->] (x2) -- (n1);
    \draw[->] (x2) -- (n2); 
    \draw[->] (x2) -- (ni); 
    \draw[->] (xn) -- (n1); 
    \draw[->] (xn) -- (n2); 
    \draw[->] (xn) -- (ni) node[midway,above] {$w_{ni}$};

\end{tikzpicture}

  \caption{Illustration of the input layer of the proposed model.}  
  \label{fig_input_lyaer_v}
\end{figure}
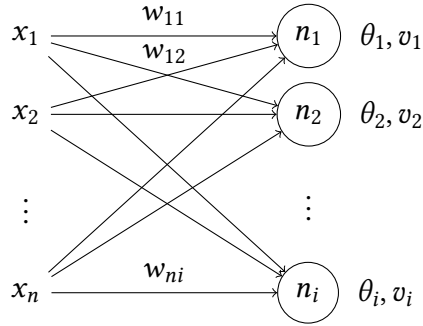

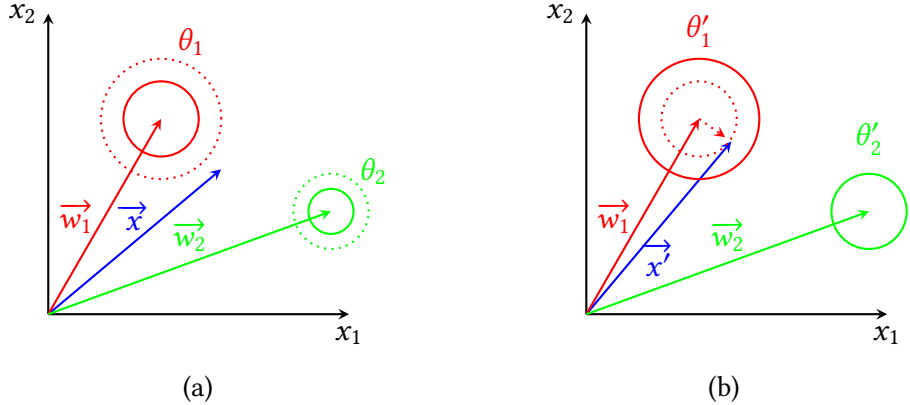
\begin{figure}[H]
\centering
\begin{subfigure}{.5\textwidth}
  \centering
\begin{tikzpicture}[>=stealth,thick]
    % Axes
    \draw[->] (0,0) -- (4,0) node[below] {$x_1$};
    \draw[->] (0,0) -- (0,4) node[left] {$x_2$};
    % Vectors
    \draw[->,blue] (0,0) -- (40:3) node[midway,above] {$\overrightarrow{x}$};
    \draw[->,red] (0,0) -- (60:3) node[midway,above, left] {$\overrightarrow{w_1}$};
    \draw[->,green] (0,0) -- (20:4) node[midway,above] {$\overrightarrow{w_2}$};
    % Circle lines
    \draw[red] (60:3) circle[radius=0.5] node[above] at (60:3.8) {$\theta_1$};
    \draw[red,dotted] (60:3) circle[radius=0.8];
    \draw[green] (20:4) circle[radius=0.3] node[above] at (20:4.6) {$\theta_2$};
    \draw[green,dotted] (20:4) circle[radius=0.5];   
\end{tikzpicture}
\caption{}
\label{fig_tresh_example_1}
\end{subfigure}~
\begin{subfigure}{.5\textwidth}
  \centering
\begin{tikzpicture}[>=stealth,thick]
    % Axes
    \draw[->] (0,0) -- (4,0) node[below] {$x_1$};
    \draw[->] (0,0) -- (0,4) node[left] {$x_2$};
    % Vectors
    \draw[->,blue] (0,0) -- (50:3) node[midway,below] {$\overrightarrow{x'}$};
    \draw[->,red] (0,0) -- (60:3) node[midway,above, left] {$\overrightarrow{w_1}$};
    \draw[->,green] (0,0) -- (20:4) node[midway,above] {$\overrightarrow{w_2}$};
    % Circle lines
    \draw[red] (60:3) circle[radius=0.8] node[above=0.8cm] {$\theta'_1$};
    \draw[red,dotted] (60:3) circle[radius=0.5];
    \draw[green] (20:4) circle[radius=0.5] node[above=0.6cm] {$\theta'_2$};
    % Red arrow between vectors
    \draw[->,red, dotted, shorten >=1mm] (60:3) -- (50:3);
\end{tikzpicture}

  \caption{}
  \label{fig_tresh_example_2}
\end{subfigure}~
\newline
\caption{An illustration of synaptic plasticity rule based on FEAST \parencite{afshar2020event} using Euclidean distance. A two-dimensional input is considered. (a) The input vector $\overrightarrow{x}$ is outside the threshold regions of both neurons. The neurons become more sensitive by increasing the thresholds. (b) A new input vector activates neuron 1. This results in adjustments of both the weights and the threshold of the active neuron.}
\label{fig_tresh_examples}
\end{figure}

The input signal is represented by a vector of real-valued observations $\overrightarrow{x}(t) = [x_1, x_2,...,x_n]$. These are provided by the environment in discrete time steps at time $t$, along with the reward signal $r(t)$. Considering $N_i$ neurons in the input layer, each neuron can be fully connected to the input signal through an array of synaptic weights, as illustrated in Figure \ref{fig_input_lyaer_v}. Euclidean distance is used as a metric to find the closest neuron $i$ with weights $\overrightarrow{w_i}$ matching the input context $\overrightarrow{x}$. Each neuron $i$ also contains a scalar threshold parameter $\theta_i$ that is used for the selection of the closest neuron. The value $v_i$ of a neuron in the input layer is calculated as the Euclidean distance between input and weight vectors:  

\begin{gather}
v_i(t) =  || \overrightarrow{w_i}(t) -  \overrightarrow{x}(t)||, \label{eq_v_update}
\end{gather}

where $v_i(t)$ is the value of the neuron $i$ at time $t$, $ \overrightarrow{w_i}$ are the weights connecting the neuron $i$ to the input vector $ \overrightarrow{x}$. The value $v_i$ of every neuron is compared to its corresponding threshold value $\theta_i$. The neuron is considered eligible if $v_i(t) < \theta_i$. A winner-take-all (WTA) activation is adopted for an entire layer or a group of $N$ eligible neurons within the same layer, so that the binary state $s_i(t)$ of each neuron is described by:
\begin{gather}
s_i(t) = \begin{cases} 
1 & \text{if } v_i(t) = \min_{j \in \{1, \ldots, N\}} v_j(t) \\
0 & \text{otherwise}
\end{cases}
\label{eq_wta}
\end{gather}
The winning neuron should have the least value among the neurons with values within their corresponding thresholds. This rule ensures that only the neuron with the least Euclidean distance within the eligible neurons will produce a spike for the next layer. If no neuron is eligible through the above rule, the neuron with the lowest value is activated but not tagged as eligible for weight adaptation. Additionally, in order to increase the probability of eligibility in the future, all neurons within the group have the threshold increased. This adaptation rule is illustrated in Figure \ref{fig_tresh_examples} and described in more detail below.

Neurons in this layer are subject to weight and threshold plasticity, with the aim of providing a compact representation of the input space. This plasticity has two drivers: i) adaptation to the input signal and ii) modulation by the feedback from the actor-critic layer. The first adaptation rule is based on Feature Extraction with Adaptive Selection Thresholds (FEAST) \parencite{afshar2020event} and on Optimized Deep Event-driven Spiking neural network Architecture (ODESA) \parencite{BethiEtAl2022}. Both FEAST and ODESA architectures use dot products to evaluate the fitness of a neuron, considering a unitary vector as an input. Meanwhile, the present work considers the input vectors to be normalized, but not unitary, i.e. real-valued observations of an RL environment and the Euclidean distance is adopted as a fitness metric.

In order to illustrate this adaptation rule, consider a two-dimensional input vector $ \overrightarrow{x} = [x_1, x_2]$, as illustrated in Figure \ref{fig_tresh_examples}. The input signal $\overrightarrow{x}$ and the weights of two input neurons are represented by two-dimensional vectors. The threshold of each neuron is a scalar and represents the receptive field of a neuron in the shape of a circle around the weight vector.  

An initial input signal $\overrightarrow{x}$, depicted in Figure \ref{fig_tresh_example_1}, falls outside the activation region of both neurons (solid circles). Thus, the value obtained by Equation \ref{eq_v_update} is higher than $\theta_1$ and $\theta_2$. When this occurs, all neurons in the layer have their thresholds increased by the parameter $\theta_\text{open}$. This increases the receptive field size of each neuron in the layer, making it more likely for a neuron to be activated by an input signal. 

On the other hand, Figure \ref{fig_tresh_example_2} illustrates the case of a new input vector $\overrightarrow{x'}$ that activates neuron 1. This is because the new value, i.e. the distance between input and weight vectors, is within the threshold of the neuron. To increase the probability of activation of the same neuron in the future, when presented with similar inputs, both the threshold and weight are adjusted. The threshold is decreased and the weight vector moves in the direction of $\overrightarrow{x'}$, illustrated by the dotted circle and arrow in Figure \ref{fig_tresh_example_2}.As a result, the receptive field of the specific winning neuron contracts, reducing the chance of activation by distant inputs.  The above adaptation, applied to activated neurons, is described by:

\begin{gather}
\Delta \theta_i = v_i(t) - \theta_i(t), \label{eq_delta_thresh}\\
\Delta \overrightarrow{w_i} =  \overrightarrow{x}(t) -  \overrightarrow{w_i}(t), \label{eq_delta_w}\\
\theta_i (t+1) = \theta_i (t) + \eta_{th} * \Delta \theta_i, \label{eq_thresh_update} \\
 \overrightarrow{w_i} (t+1) =  \overrightarrow{w_i} (t) + \eta * \Delta  \overrightarrow{w_i}, \label{eq_w_update}
\end{gather}
where $\Delta \theta_i$ and $\Delta  \overrightarrow{w_i}$ are update values for the threshold and weights of neuron $i$ and the respective update rates for each are $\eta_{th}$ and $\eta$.

\begin{figure}[H]
\centering
\begin{subfigure}{.5\textwidth}
  \centering
  \includegraphics[height=2.0 in]{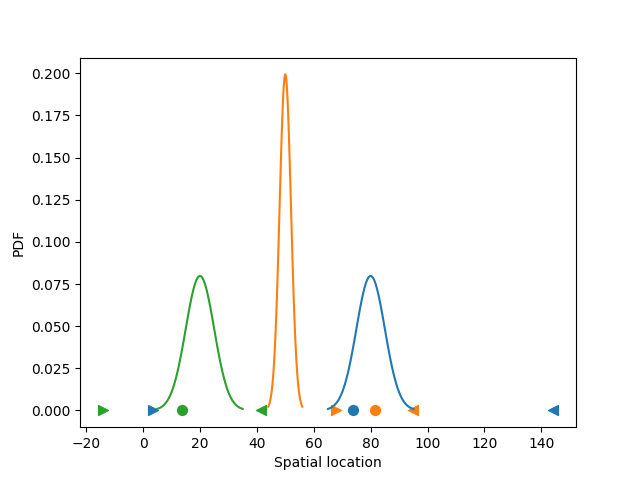}
  \caption{Initial position}
  \label{fig_sample_1d_clusters_before}
\end{subfigure}~
\begin{subfigure}{.5\textwidth}
  \centering
  \includegraphics[height=2.0 in]{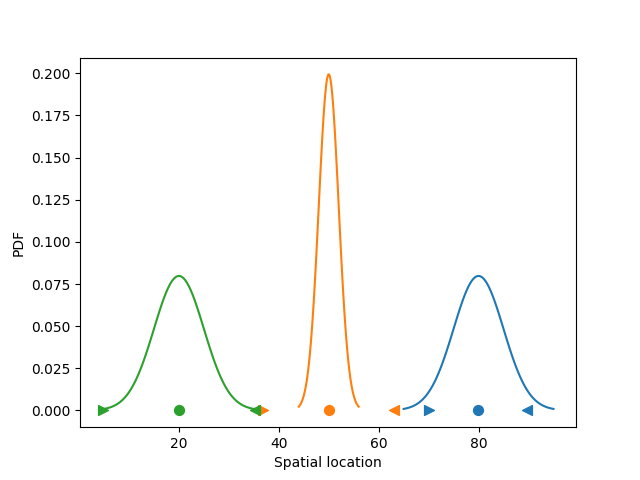}
  \caption{After adaptation}
  \label{fig_sample_1d_clusters_after}
\end{subfigure}~
\newline
\caption{Clustering of neurons in the input layer on a simulated dataset with three Gaussian distributions and three input neurons. The weights of the three neurons are depicted by the colored dots and the corresponding thresholds are represented by the same-colored arrows.}
\label{fig_sample_1d_clusters}
\end{figure}

Figure \ref{fig_sample_1d_clusters} illustrates this adaptation process on a simulated dataset with three probability density functions (PDF). Three neurons are initialized with random weights and thresholds, as illustrated in Figure \ref{fig_sample_1d_clusters_before}. The weight of each neuron is depicted as a colored dot and the threshold region is delimited by same-colored arrows. The final configuration is depicted in Figure \ref{fig_sample_1d_clusters_after}, after $10^4$ observations and using both $\eta_{th}$ and $\eta$ equal to $10^{-2}$.

\begin{figure}[H]
\centering
\begin{subfigure}{.5\textwidth}
  \centering
  \includegraphics[height=2.0 in]{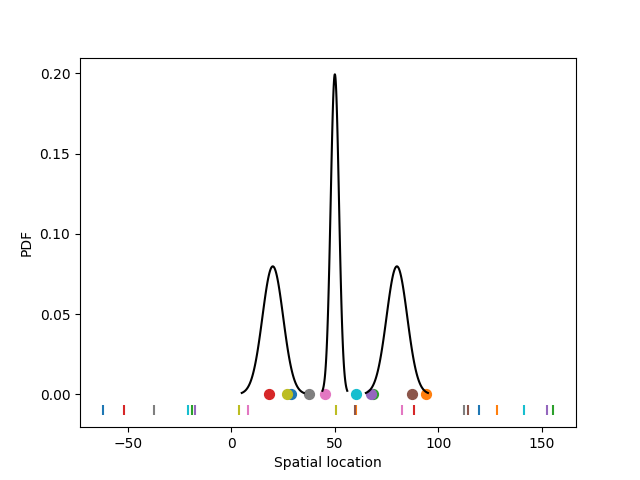}
  \caption{Initial position}
  \label{fig_sample_1d_10_clusters_before}
\end{subfigure}~
\begin{subfigure}{.5\textwidth}
  \centering
  \includegraphics[height=2.0 in]{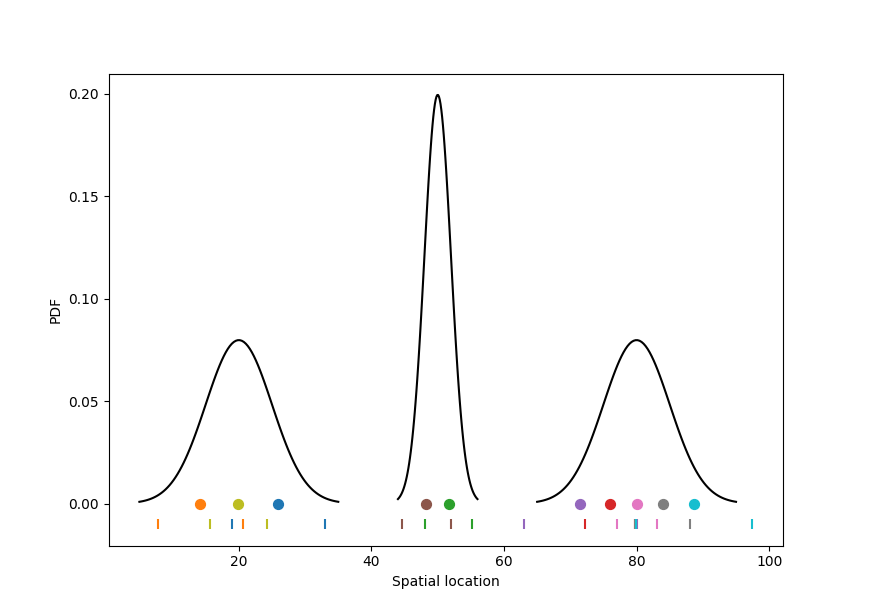}
  \caption{After adaptation}
  \label{fig_sample_1d_10_clusters_after}
\end{subfigure}~
\newline
\caption{Clustering of neurons in the input layer on a simulated dataset with three Gaussian distributions and ten input neurons. The positions of the colored dots on the X-axis represent the weights of the ten neurons used in the experiment. For better legibility, thresholds are represented as vertical lines of the same color below the dots.}
\label{fig_sample_1d_10_clusters}
\end{figure}

Note that if more neurons are added to the clustering layer, the adaptation process described by Equations \ref{eq_delta_thresh} to \ref{eq_w_update} will result in a denser, more granular representation of the observed state-space via neurons with smaller thresholds. This is experimentally demonstrated in Figures \ref{fig_sample_1d_10_clusters_before} and \ref{fig_sample_1d_10_clusters_after} by using the same set of parameters as in the previous experiments but increasing the number of neurons from 3 to 10.

In addition to the weight and threshold plasticity rule above, the proposed model implements modulation by feedback from the actor-critic layer. This is similar to the change in Equations \ref{eq_thresh_update} and \ref{eq_w_update}, but $\eta_{th}$ and $\eta$ are scaled by temporal difference error $\delta$ and neural trace $c_i$. This modulation is described in more detail in Section \ref{ssec_proposed_SNN_ac_layer}.

\begin{figure}[H]
\centering
\begin{subfigure}{.4\textwidth}
  \centering
  \resizebox{.35\columnwidth}{!}{%
    
   \begin{tikzpicture}[node distance=.5cm]
    % Nodes y11, y12, y13, and y14
    \node[draw, circle] (y11) {};
    \node[draw, circle, fill=black, below=of y11] (y12) {};
    \node[below=of y12] (y13) {\vdots};
    \node[draw, circle, below=of y13] (y14) {};

    % First input signal
    \node[below left=0.5cm and 1cm of y11] (x1) {$x_1$};

    \node[draw, dotted, fit=(y11) (y14), inner sep=5pt] (n1) {};

    \draw (x1) -- (y11);
    \draw (x1) -- (y12);
    \draw (x1) -- (y14);

    % Last input signal
    \node[below=1cm of x1] (xn) {$x_n$};

    \draw (xn) -- (y11);
    \draw (xn) -- (y12);
    \draw (xn) -- (y14);

    % Position of the dots
    \path (x1) -- node[midway] (dots) {\vdots} (xn);
    \end{tikzpicture}

  }
  \caption{}
  \label{fig_input_layer_conf_1}
\end{subfigure}~
\begin{subfigure}{.4\textwidth}
  \centering
  \resizebox{.3\columnwidth}{!}{
    \begin{tikzpicture}[node distance=.3cm]
         \node (dots) at (0,0) {\vdots};

         % First input signal
         \node[above left=1.25cm and 0cm of dots] (x1) {$x_1$};
         \node[draw, circle, fill=black, above right=0cm and 1cm of x1] (y12) {};
         \node[draw, circle, above=of y12] (y11) {};
         \node[below right=0cm and 1cm of x1] (y13) {\vdots};
         \node[draw, circle, below=of y13] (y14) {};

         \node[draw, dotted, fit=(y11) (y14), inner sep=4pt] (n1) {};

         \draw (x1) -- (y11);
         \draw (x1) -- (y12);
         \draw (x1) -- (y14);

         %\node[draw, dashed, rounded corners, fit=(x1) (n1), inner sep=10pt] (box1) {};

         % Last input signal
         \node[below left=1.1cm and 0cm of dots] (xn) {$x_n$};
         \node[draw, circle, above right=0cm and 1cm of xn] (yn2) {};
         \node[draw, circle, above=of yn2] (yn1) {};
         \node[below right=0cm and 1cm of xn] (yn3) {\vdots};
         \node[draw, circle, fill=black, below=of yn3] (yn4) {};

         \node[draw, dotted, fit=(yn1) (yn4), inner sep=4pt] (yboxn) {};

         \draw (xn) -- (yn1);
         \draw (xn) -- (yn2);
         \draw (xn) -- (yn4);

        % \node[draw, rounded corners, fit=(xn) (yboxn), inner sep=10pt] (boxn) {};
     \end{tikzpicture}
    }
  \caption{}
  \label{fig_input_layer_conf_2}
\end{subfigure}~
\newline
\caption{Diagrams of evaluated configurations for encoding the input vector by the first layer. (a) The input is fully connected to the neurons in the first layer with WTA activation. (b) A group of neurons if fully connected to a single input scalar and WTA is applied to each group separately.}
\label{fig_input_layer_confs}
\end{figure}
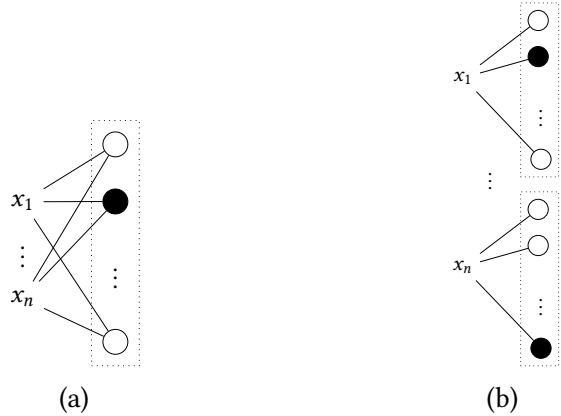

The input layer neurons can encode the real-valued vector $\overrightarrow{x}$ in a number of ways. In the present work, we consider two possible configurations, as depicted in Figure \ref{fig_input_layer_confs}: a) fully connected and b) individual encoding for each input dimension, followed by an additional clustering layer. Neurons grouped within a dotted rectangle represent a winner-take-all (WTA) activation mechanism, where the neuron possessing the highest potential transmits a pulse to the succeeding layer.

Note that the configuration in Figure \ref{fig_input_layer_conf_2} produces multiple active neurons at each time-step, corresponding to the size of the input vector $\overrightarrow{x} = [x_1, x_2,...,x_n]$. In this case, a hidden clustering layer is used to provide a single winner neuron for the actor-critic layer, as illustrated in Figure \ref{fig_snn_overview}. The WTA activation permits only one neuron from this second layer to generate a spike at any time step, thus explicitly enforcing sparsity at the layer level. The synapses between the first and second clustering layers are subject to the same modulation, as described by Equations \ref{eq_delta_thresh} to \ref{eq_w_update}, replacing the input vector $\overrightarrow{x}$ by a larger multi-hot binary vector, representing activations in the previous layer.

\subsection{Actor-critic layer}
\label{ssec_proposed_SNN_ac_layer}

The actor-critic layer is a neuromorphic implementation of the tabular TD$(\lambda)$ algorithm \parencite{sutton1988learning}. The actor is composed of $N_a$ neurons, each representing a single discrete action. When a spike is produced by a single presynaptic neuron $j$, the resulting potential of an action neuron $i$ is equal to the synaptic value $w_{ij}$. The action is then selected based on the highest potential, with the addition of random exploration:

\begin{gather}
a(t) = \begin{cases}
\underset{i}{\arg\max}\ w_{i,j}, & \text{with  probability $1-\epsilon$}\\
    \textit{random}, & \text{with probability $\epsilon$},
  \end{cases}
\label{eq_action_selection}
\end{gather}

where $\epsilon$ is the exploration probability. At the beginning of an episode $\epsilon$ is initialized at a maximum value of 1 and is then gradually decreased to a small final value $\epsilon_{min}$. The rate of decrease and the final baseline value are hyperparameters used to balance exploration and exploitation. 

In the proposed system an action can be the result of a single or multiple neurons in the action layer spiking by using WTA activation. For instance, a movement in two dimensions can be encoded by two independent groups of neurons, each representing a one-dimensional action.  

The value of a hidden state is represented by the weight connecting the value neuron to the hidden layer. Two separate synaptic eligibility traces are implemented for the actor and critic connections. These traces decay exponentially over time:

\begin{gather}
\overrightarrow{c_c}(t+1) = \overrightarrow{c_c}(t) - \frac{\overrightarrow{c_c}(t)}{\tau_c}, \label{eq_trace_decay_critic}
\\
\overrightarrow{c_a}_i(t+1) = \overrightarrow{c_a}_i(t) - \frac{\overrightarrow{c_a}_i(t)}{\tau_a},\label{eq_trace_decay_actor}
\end{gather}

where $\tau_{a}$ and $\tau_{c}$ are time constants for the actor and critic traces respectively. Correspondingly, actor and critic trace vectors are denoted as $\overrightarrow{c_a}_i$ and $\overrightarrow{c_c}$. Note that $\overrightarrow{c_a}_i$ relates to the action neuron $i$ and $c_{aij}$ is set to one when an action neuron $i$ and hidden neuron $j$ fire at the same discrete time step.

Finally, after an action is selected, the environment provides the next observation $\overrightarrow{x}(t+1)$ and reward signal $r(t+1)$. Based on this feedback, the temporal difference (TD) error is calculated as follows \parencite{sutton2018reinforcement}: 

\begin{gather}
\delta = r(t+1) + \gamma V(t+1) - V(t), \label{eq_td_error}
\end{gather}
where $V(t)$ is the value of the hidden state at time $t$ and $\gamma$ is a discount parameter. Based on the TD error, value and action weights are updated:

\begin{gather}
\overrightarrow{w_v}(t+1) = \overrightarrow{w_v}(t) + \overrightarrow{c_c}(t) * \eta_c * \delta, \label{eq_update_value} \\
\overrightarrow{w_a}_i(t+1) = \overrightarrow{w_a}_i(t) + \overrightarrow{c_a}_i(t) * \eta_a * \delta, \label{eq_update_act}
\end{gather}

where $\overrightarrow{w_v}$ is a vector of values, $\overrightarrow{w_a}_i$ are weights arriving at the action neuron $i$. Constants $\eta_c$ and $\eta_a$ are update rates for the critic and actor respectively. Finally, weights and thresholds of clustering layers are also modulated by the TD error:

\begin{gather}
\theta_i(t+1) = \theta_i(t) + \eta_{td} * \Delta \theta_i * |\delta| * c_i(t), \label{eq_thresh_td_update} \\
 \overrightarrow{w_i}(t+1) =  \overrightarrow{w_i}(t) + \eta_{td} * \Delta  \overrightarrow{w_i} * |\delta| * c_i(t), \label{eq_w_td_update}
\end{gather}
where $\eta_{td}$ is the update rate and $c_i(t)$ is the activation trace of the postsynaptic neuron $i$. Following Equation \ref{eq_trace_decay_actor}, this trace is set to one each time neuron $i$ is eligible and decays exponentially with time constant $\tau_{i}$. This update occurs concurrently with unsupervised clustering described by Equations \ref{eq_thresh_update} and \ref{eq_w_update}.

With the above update rule lower absolute TD error causes the neuron to follow the unsupervised clustering rule from Equations \ref{eq_thresh_update} and \ref{eq_w_update} and illustrated in Figure \ref{fig_tresh_examples}. Conversely, a higher positive or negative TD error signal makes the model more likely to retain the recent feature representation. This behaviour causes neurons to become more sensitive to regions of the input space that correspond to higher absolute TD errors.  

\begin{figure}[H]
\centering
\begin{subfigure}{.9\textwidth}
  \centering
  \includegraphics[width=.8\textwidth]{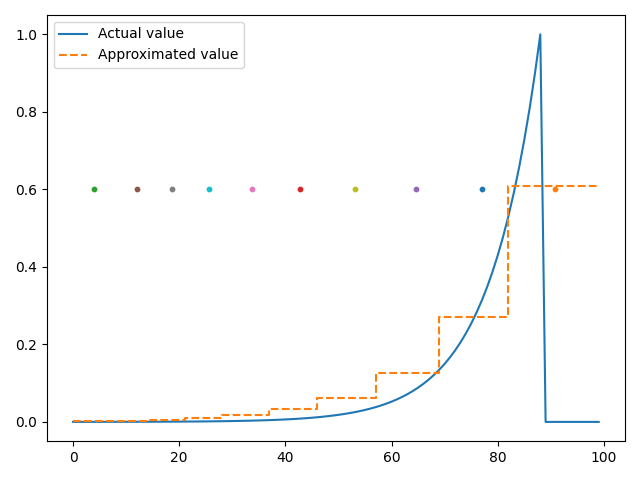}
  \caption{}
  \label{fig_clustering_linear_track_no_modulation}
\end{subfigure}~ 
\newline
\begin{subfigure}{.9\textwidth}
  \centering
  \includegraphics[width=.8\textwidth]{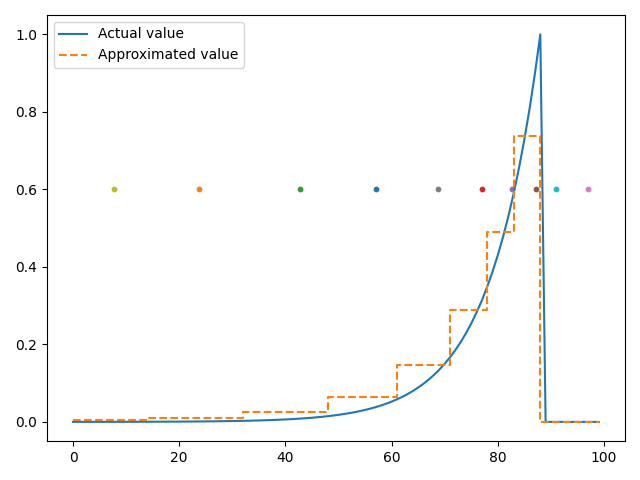}
  \caption{}
  \label{fig_clustering_linear_track_tdm}
\end{subfigure}~
\newline
\caption{An illustration of the role of TD-modulated clustering for a more accurate value representation. Position of the colored dots on the X-axis represent the weights of the ten neurons used in the experiment. Thresholds are omitted for legibility.}
\label{fig_clustering_linear_track}
\end{figure}

A linear track experiment is used to illustrate the role of TD-modulated clustering. In this environment, the agent travels through a fixed track from position 1 to 100 and each time-step moves the agent to the right by one unit. A single reward of +1 is provided at position 90 and once the end of the track is reached, the agent restarts at position 1. Considering a discount $\gamma=0.9$, the value corresponding to each position is represented by a solid blue line in Figure \ref{fig_clustering_linear_track}. Since in the proposed architecture, the state values are estimated from a hidden layer, a discrete approximation of a real-valued curve is formed by the ten hidden neurons. This approximation is represented by a dotted orange line and the solid dots correspond to the weights of these hidden neurons after 500 episodes. Note that if no TD-modulation is used, the weights would converge to a uniform representation of the observed state-space, as depicted in Figure \ref{fig_clustering_linear_track_no_modulation}. On the other hand, by adding modulation through TD-error, the weights will be skewed towards the states that produce the highest TD-error, which is illustrated by a more accurate approximation of the value curve in Figure \ref{fig_clustering_linear_track_tdm}. 

\begin{figure}[H]
\centering
  \includegraphics[width= \linewidth]{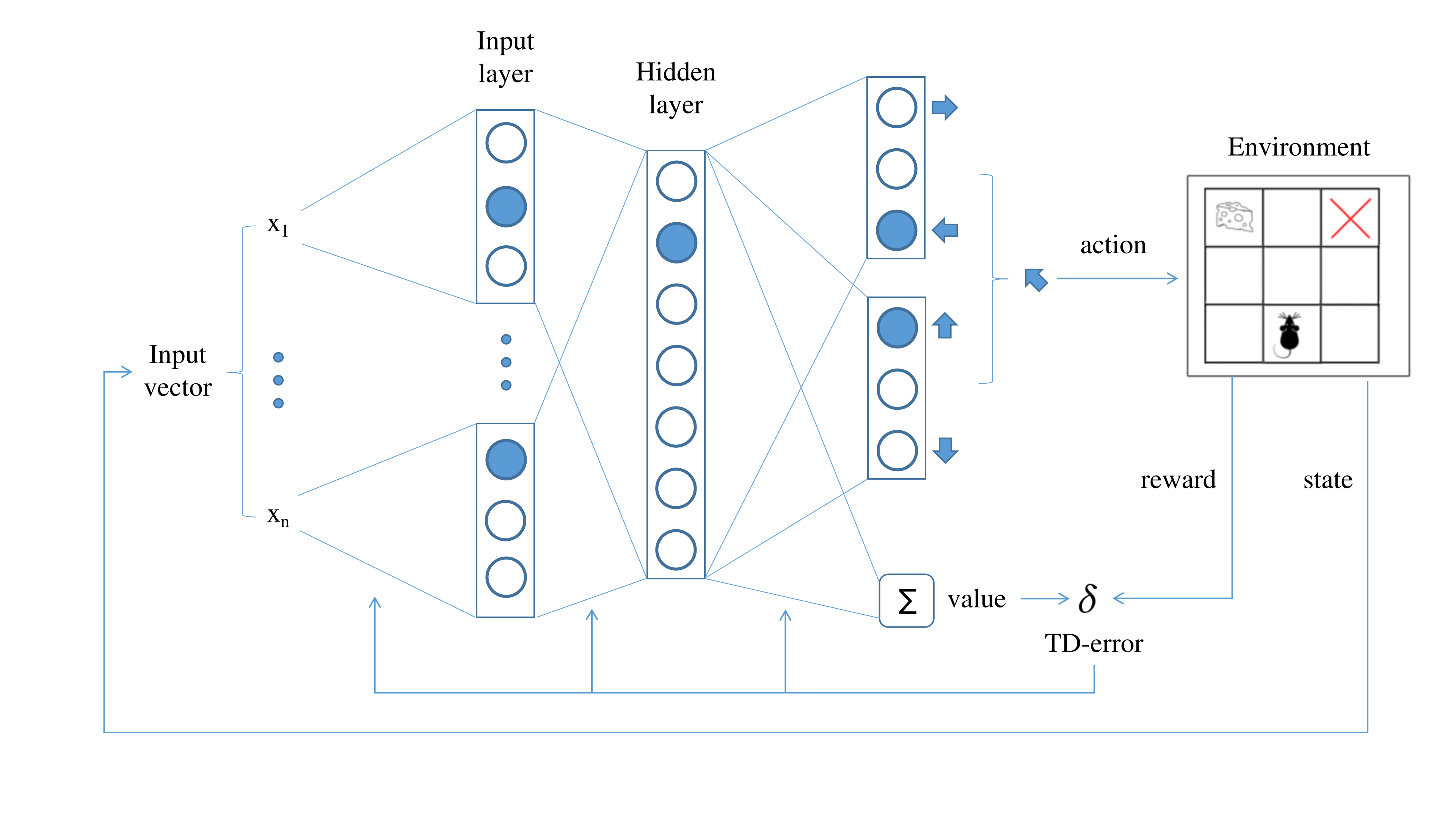}  
  \caption{An overview of the presented network. In this example, each input scalar $x_i$ is connected to a group of three neurons in the first clustering layer. A WTA activation is applied independently on each group and a resulting multi-hot vector is processed further by the hidden clustering layer. Another WTA activation function, this time applied to the entire layer, results in a single active neuron which represents a distinct hidden state. Orthogonal actions are independently encoded by subsequent groups of neurons and the value corresponding to the hidden state is computed based on sum of the synaptic weights. Eligibility traces are attached to both neurons and synapses for weight and threshold modulation.}  
  \label{fig_snn_overview}
\end{figure}

An overview of the proposed network is illustrated in Figure \ref{fig_snn_overview}. In this example, the first clustering layer is divided into groups of three neurons, each with a WTA activation. The activation trace from this first layer is a higher dimensional vector than the input. Both the first and second layers implement the FEAST clustering rule proposed by \Textcite{afshar2020event}, as well as additional modulation by TD error, described by Equations \ref{eq_thresh_td_update} and \ref{eq_w_td_update}.

\section{Experimental evaluation}
\label{sec_exp_eval}

The proposed architecture is experimentally compared to a classical tabular actor-critic algorithm with eligibility traces (TAC), as well as a state-of-the-art deep reinforcement learning (DRL) algorithm PPO. Three classic control RL environments from OpenAI Gym \parencite{brockman2016openai} are used for this evaluation: mountain car, cart-pole and acrobot. The sections below provide a more detailed description of these environments and baseline algorithms.

\subsection{RL environments} \label{sec_exp_environments}

\subsubsection{Mountain car} \label{ssec_exp_environments_mountain_car}

\begin{figure}[H]
\centering
  \includegraphics[width= 0.5 \linewidth]{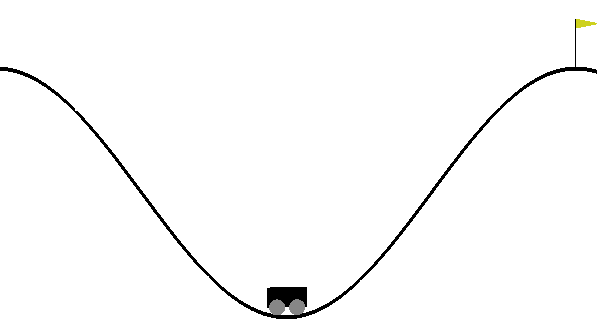}  
  \caption{The mountain car environment \parencite{brockman2016openai}.}  
  \label{fig_env_mountain_car}
\end{figure}

The Mountain car is a classic RL control problem, illustrated in Figure \ref{fig_env_mountain_car}. The environment consists of a car that is stuck in a valley between two hills. The car has limited power and is unable to climb the hills directly. The goal of the agent is to learn how to control the car so that it can reach the top of the hill on the right.

This environment has a two-dimensional state space, consisting of the position and velocity of the car. The position is a continuous value that ranges from -1.2 to 0.6, while the velocity ranges from -0.07 to 0.07. The action space is discrete, consisting of three possible actions: push the car to the left, push the car to the right, or do nothing.

The reward function in the Mountain car environment is designed to encourage the agent to reach the top of the hill on the right. The agent receives a reward of -1 at each time step until it reaches the goal. This penalization is added to encourage the agent to reach the goal as quickly as possible. Once the agent reaches the goal, the episode ends and the agent receives a reward of 0. 

This is a non-trivial problem because of the long-term dependencies involved. The agent must learn to build up momentum by moving back and forth in the valley before it can climb the hill on the right. This requires the agent to balance short-term rewards with long-term goals. Importantly, the goal state is extremely unlikely to be reached by chance, i.e. random exploration, within the time limit of 200 steps.

\subsubsection{Cart-pole} \label{ssec_exp_environments_Cart-pole}

\begin{figure}[H]
\centering
  \includegraphics[width= 0.5 \linewidth]{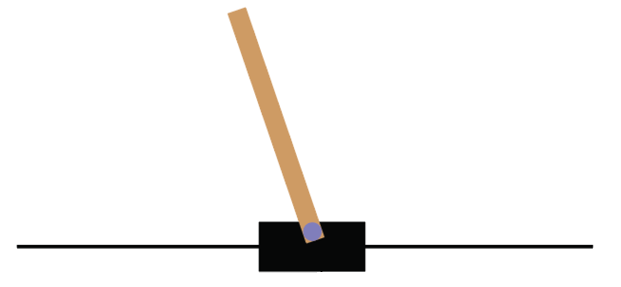}  
  \caption{The Cart-pole environment \parencite{brockman2016openai}.}  
  \label{fig_env_cart_pole}
\end{figure}

The cart-pole environment in OpenAI Gym \parencite{brockman2016openai} is another classic RL problem, illustrated in Figure \ref{fig_env_cart_pole}. The environment consists of a cart and a pole that is attached to the cart by a free-moving joint. The goal of the agent is to balance the pole on top of the cart for as long as possible.

The state space of the Cart-pole environment is four-dimensional, consisting of the position and velocity of the cart, and the angle and angular velocity of the pole. The cart position and pole angle are continuous values, as well as the respective velocities. The action space is discrete, consisting of two possible actions: move the cart to the left or move the cart to the right.

The reward function in the Cart-pole environment is designed to encourage the agent to balance the pole on top of the cart for as long as possible, limited to 500 steps. The agent receives a reward of +1 at each time step while the pole remains upright. The episode ends when the pole falls over or when the cart moves too far to the left or right. Once the episode ends, the agent receives a reward of 0. 

The Cart-pole is a challenging RL problem because of its instability. The agent must learn to balance the pole while also avoiding moving the cart too far to either side. 

\subsubsection{Acrobot} \label{ssec_exp_environments_acrobot}
\begin{figure}[H]
\centering
  \includegraphics[width= 0.5 \linewidth]{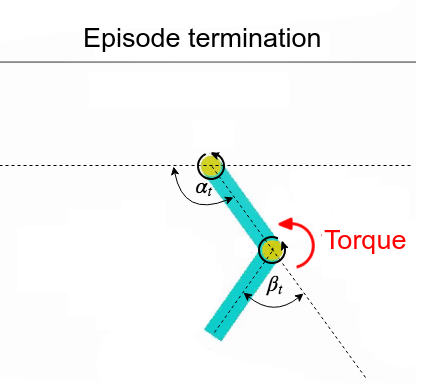}  
  \caption{The acrobot environment \parencite{brockman2016openai}.}  
  \label{fig_env_acrobot}
\end{figure}

 The setup is illustrated in Figure \ref{fig_env_acrobot} and the goal is to lift the tip of the robot to a target level. The environment provides a six-dimensional observation consisting of the sine and cosine of both joint angles, as well as the respective angular velocities. The second joint is weakly actuated and the system includes gravitational pull. To solve this task, the agent has to consistently swing the actuated joint, building up the energy.
 
The reward function in the Acrobot environment is designed to encourage the agent to swing the pendulum up to a vertical position. The agent receives a reward of -1 at each time step, as with the other environments, this encourages the agent to reach the goal as quickly as possible. When the pendulum reaches the goal position the episode ends and the environment provides a reward of 0. 

\subsection{Baseline algorithms} \label{sec_exp_baselines}

\subsubsection{TAC} \label{ssec_exp_baselines_td_lambda}

The tabular actor-critic algorithm is based on a classical TD($\lambda$) reinforcement learning algorithm used to estimate the optimal value function of an MDP, given a set of states, actions, and rewards \parencite{sutton1988learning}. It is a variant of the TD-learning algorithm that incorporates eligibility traces, which allow for the accumulation of TD-errors over multiple time steps.

At each time step, the algorithm updates the value function estimate of the current state $s$ using the TD-error $\delta$, defined in Equation \ref{eq_td_error}. This update is weighted by the eligibility trace, which is a measure of the importance of the state $s$ in the current episode.

The eligibility traces of the actor and critic tables are updated at each time step using the decay parameters $\tau_a$ and $\tau_c$, as well as the discount factor $\gamma$. This allows the algorithm to incorporate information from previous time steps into the current update and improve the accuracy of the value function estimate.

The algorithm repeats this process for a fixed number of episodes, updating the value function estimate and eligibility traces at each time step. The epsilon-greedy exploration strategy is used to balance exploration and exploitation.

\subsubsection{PPO} \label{ssec_exp_baselines_PPO}

Proposed by \Textcite{schulman2017proximal}, Proximal Policy Optimization (PPO) is a RL algorithm used for training policy-based models, which directly optimizes the policy function that maps states to actions. It is a variant of policy gradient methods, which aim to maximize the expected reward of an agent by adjusting the parameters of the policy function.

PPO uses a surrogate objective function that constrains the change in the policy parameters to be small so that the agent does not deviate too far from the previous policy. This makes the training process less brittle and prevents the agent from making drastic policy changes. The PPO algorithm uses a clipping technique to enforce this constraint on the policy updates. At each iteration, the algorithm computes the ratio between the new and old policy probabilities for the observed state-action pairs and uses this ratio to compute a surrogate objective function. The surrogate objective function is then optimized using a gradient descent algorithm, subject to a constraint that limits the size of the policy update.

The size of the policy update is controlled by a hyperparameter $\epsilon_{\text{clip}}$, which represents the maximum amount that the policy parameters can change in a single update. If the ratio of the new and old policy probabilities exceeds $1 + \epsilon_{\text{clip}}$, then the update is clipped to a maximum value of $1 + \epsilon_{\text{clip}}$. Similarly, if the ratio is less than $1 - \epsilon_{\text{clip}}$, then the update is clipped to a minimum value of $1 - \epsilon_{\text{clip}}$. This helps to prevent the policy from changing too much and ensures that the updates are consistent with the previous policy.

Analogously to the actor-critic model in the proposed network,  PPO also uses a value function to estimate the expected reward of each state.

\subsection{Hyperparameters} \label{sec_exp_hyperparams}
This section presents the main hyperparameters that were found to have an impact on each algorithm's performance during exploratory trials. A widely used  optimization framework \textit{Optuna}~\parencite{optuna_2019} is employed to iteratively search for a good set of hyperparameters for the proposed network and both of the baseline algorithms. More specifically, the Tree of Parzen estimators (TPE) \cite{bergstra2011algorithms} algorithm is used for sequential optimization. After an initial random exploration for 10 trials, TPE aims at choosing a set of parameters that maximize an objective function. If a model with the same set of parameters is evaluated more than once, the final score is an average of all the trials within this set.  

\subsubsection{TAC} \label{ssec_exp_hyperparams_td_lambda}
The tabular actor-critic baseline is optimized for 500 trials on each of the three benchmark problems: mountain car, cart-pole and acrobot. Each trial is composed of 1000 episodes and the optimizer aims at achieving the best average latency at the last 500 episodes. The best-performing parameters for each of these environments are indicated in Table \ref{tab_hyper_td_lambda} with letters \textbf{a}, \textbf{b} and \textbf{c} corresponding to mountain car, cart-pole and acrobot, respectively. 

\begin{table}[H]
%\setlength\tabcolsep{5pt}
%\def\arraystretch{1.2}
%\scriptsize
\centering
\caption{Hyperparameter search space for the TAC baseline. Letters indicate the selected configuration for each environment: mountain car (\textbf{a}), cart-pole (\textbf{b}) and acrobot (\textbf{c}).}
\label{tab_hyper_td_lambda}
\begin{tabular}{lcc}
\hline
\multicolumn{1}{c}{\textbf{Hyperparameter}} & \textbf{Search space}\\ \hline \hline
\# of bins per dimension & [5, 10(\textbf{b},\textbf{c}), 20(\textbf{a})]\\
$\epsilon_{min}$ & [0.01(\textbf{a}), 0.05(\textbf{b},\textbf{c}), 0.1]\\
$\epsilon$ decay time (\# of episodes)  & [100, 200(\textbf{b},\textbf{c}), 500(\textbf{a})]\\
Discount factor & [0.9(\textbf{c}), 0.95(\textbf{a},\textbf{b}), 0.99]\\
Actor learning rate & [$10^{-3}$, $10^{-2}$(\textbf{b},\textbf{c}), $10^{-1}$(\textbf{a})]\\
Critic learning rate & [$10^{-3}$, $10^{-2}$(\textbf{b},\textbf{c}), $10^{-1}$(\textbf{a})]\\

$\tau_{a}$ & [1(\textbf{a},\textbf{b},\textbf{c}), 10, 20]\\
$\tau_{c}$ & [1, 10, 20(\textbf{a},\textbf{b},\textbf{c})]\\
\hline
\end{tabular}
\end{table}

\subsubsection{PPO} \label{ssec_exp_hyperparams_PPO}
Table \ref{tab_hyper_ppo} presents the list of hyperparameters for the PPO algorithm, described in Section \ref{ssec_exp_baselines_PPO}. The optimization process is identical to the TAC algorithm. Due to the sparse rewards of the mountain car problem, the PPO algorithm was not used on this environment.   

\begin{table}[H]
%\setlength\tabcolsep{5pt}
%\def\arraystretch{1.2}
%\scriptsize
\centering
\caption{Hyperparameter search space for the PPO baseline. Letters indicate the selected configuration for each environment: mountain car (\textbf{a}), cart-pole (\textbf{b}) and acrobot (\textbf{c}).}
\label{tab_hyper_ppo}
\begin{tabular}{lcc}
\hline
\multicolumn{1}{c}{\textbf{Hyperparameter}}                                 & \textbf{Search space}\\ \hline \hline
\# of neurons in each hidden layer & [64(\textbf{b},\textbf{c}), 128, 256]\\
\# of hidden layers & [1(\textbf{b}), 2(\textbf{c}), 3] \\
Buffer size & [1k, 2k(\textbf{b},\textbf{c}), 4k]\\
Learning rate & [$10^{-3}$, $10^{-4}$(\textbf{b},\textbf{c}), $10^{-5}$]\\
Batch size & [32, 64(\textbf{b},\textbf{c}), 128]\\
Discount factor & [0.9(\textbf{c}), 0.95,  0.99(\textbf{b})]\\
\# of surrogate loss optimization episodes & [5, 10(\textbf{b},\textbf{c}), 20]\\
Clipping parameter $\epsilon_{\text{clip}}$ & [0.1, 0.2(\textbf{b},\textbf{c}), 0.4]\\
\hline
\end{tabular}
\end{table}

\subsubsection{Proposed model} \label{ssec_exp_hyperparams_proposed}
The proposed model shares hyperparameters with the TAC algorithm, as the Actor-Critic part of the network is a neuromorphic implementation of this algorithm. However, a different search space is selected because the proposed Actor-Critic model does not evaluate the entire state-space of a given environment. Rather, a hidden layer of the clustering network is used to partition the state-space into a more compact and discrete representation. 

In contrast to the baseline algorithms above, and due to the large number of adjustable parameters in the proposed model, a hybrid optimization strategy was adopted. The TPE algorithm was used alternatively with manual tuning, typically adjusting a few parameters at a time. The final set of optimized parameters is presented in Table \ref{tab_hyper_proposed}, with letters \textbf{a}, \textbf{b} and \textbf{c} indicating the hyperparameters found for mountain cat, cart-pole and acrobot environments. Note that only the acrobot environment benefited from a second clustering layer. Thus, an additional optimized configuration with a single clustering layer is indicated by \textbf{d}. This configuration is used to evaluate the impact of the second clustering layer, as described in Section \ref{sssec_discussion}.

\begin{table}[H]
%\setlength\tabcolsep{5pt} 
%\def\arraystretch{1.2}
%\scriptsize
\centering
\caption{Hyperparameter search space for the proposed network. Letters indicate the selected configuration for each environment: mountain car (\textbf{a}), cart-pole (\textbf{b}) and acrobot (\textbf{c} and \textbf{d}).}
\label{tab_hyper_proposed}
\begin{tabular}{lc}
\hline
\multicolumn{1}{c}{\textbf{Hyperparameter}} & \textbf{Search space}\\ \hline \hline
$\eta$ & [$10^{-5}$(\textbf{a}), $10^{-4}$, $10^{-3}$(\textbf{b},\textbf{c},\textbf{d}), $10^{-2}$]\\
$\eta$ decay factor & [$10^{-3}$(\textbf{b}), $10^{-2}$(\textbf{d}), $10^{-1}$(\textbf{a},\textbf{c}), $1$]\\
$\eta$ decay time (\# of episodes)  & [100(\textbf{b},\textbf{d}), 500(\textbf{c}), 1000(\textbf{a})]\\

% threshold_open
$\theta_{\text{open}}$ & [$10^{-3}$(\textbf{b}), $10^{-2}$(\textbf{a},\textbf{c}), $10^{-1}$(\textbf{d})]\\
$\theta_{\text{open}}$ decay factor & [$10^{-3}$(\textbf{b}), $10^{-2}$, $10^{-1}$(\textbf{a},\textbf{c},\textbf{d}), $1$]\\
$\theta_{\text{open}}$ decay time (\# of episodes)  & [100(\textbf{b},\textbf{d}), 500, 1000(\textbf{a},\textbf{c})]\\

\# of neurons in the first clustering layer & [10, 20(\textbf{c}), 50, 100(\textbf{a},\textbf{b},\textbf{d})]\\
\# of neurons in the second clustering layer & [10, 20(\textbf{c}), 50]\\

% eligibility_change_rate
$\eta_{td}$ & [$10^{-3}$(\textbf{a}), $10^{-2}$(\textbf{c},\textbf{d}), $10^{-1}$(\textbf{b})]\\

% actor-critic parameters:
$\epsilon_{min}$ & [0.01(\textbf{a},\textbf{b},\textbf{c},\textbf{d}), 0.05, 0.1]\\
$\epsilon$ decay time (\# of episodes)  & [100, 200, 500(\textbf{a},\textbf{b},\textbf{c},\textbf{d})]\\
Discount factor & [0.9, 0.95(\textbf{b},\textbf{c}), 0.99(\textbf{a},\textbf{d})]\\
Actor learning rate & [$10^{-3}$, $10^{-2}$, $10^{-1}$(\textbf{a},\textbf{b},\textbf{c},\textbf{d})]\\
Critic learning rate & [$10^{-3}$, $10^{-2}$(\textbf{c}), $10^{-1}$(\textbf{a},\textbf{b},\textbf{d})]\\
$\tau_{a}$ & [1, 5(\textbf{a}), 10(\textbf{c},\textbf{d}), 20, 50(\textbf{b})]\\
$\tau_{c}$ & [1, 5(\textbf{a}), 10(\textbf{b},\textbf{c},\textbf{d}), 20, 50]\\
\hline
\end{tabular}
\end{table}

\subsection{Results}
\label{sec_exp_results}

The following experimental results are obtained from ten independent runs for each combination of an RL algorithm and a benchmark environment. A random agent ($\epsilon=1$) is also included in the experiments. The results are presented as the average of the ten trials, with respect to each episode. Additionally, the trial with the best latency is shown separately as a dotted line and the shaded region indicates standard deviation, also relative to the episode. 
%\subsubsection{Grid world} \label{ssec_exp_results_grid_world}

\subsubsection{Mountain car} \label{ssec_exp_results_mountain_car}

\begin{figure}[H]
\centering
  \includegraphics[width= 0.8 \linewidth]{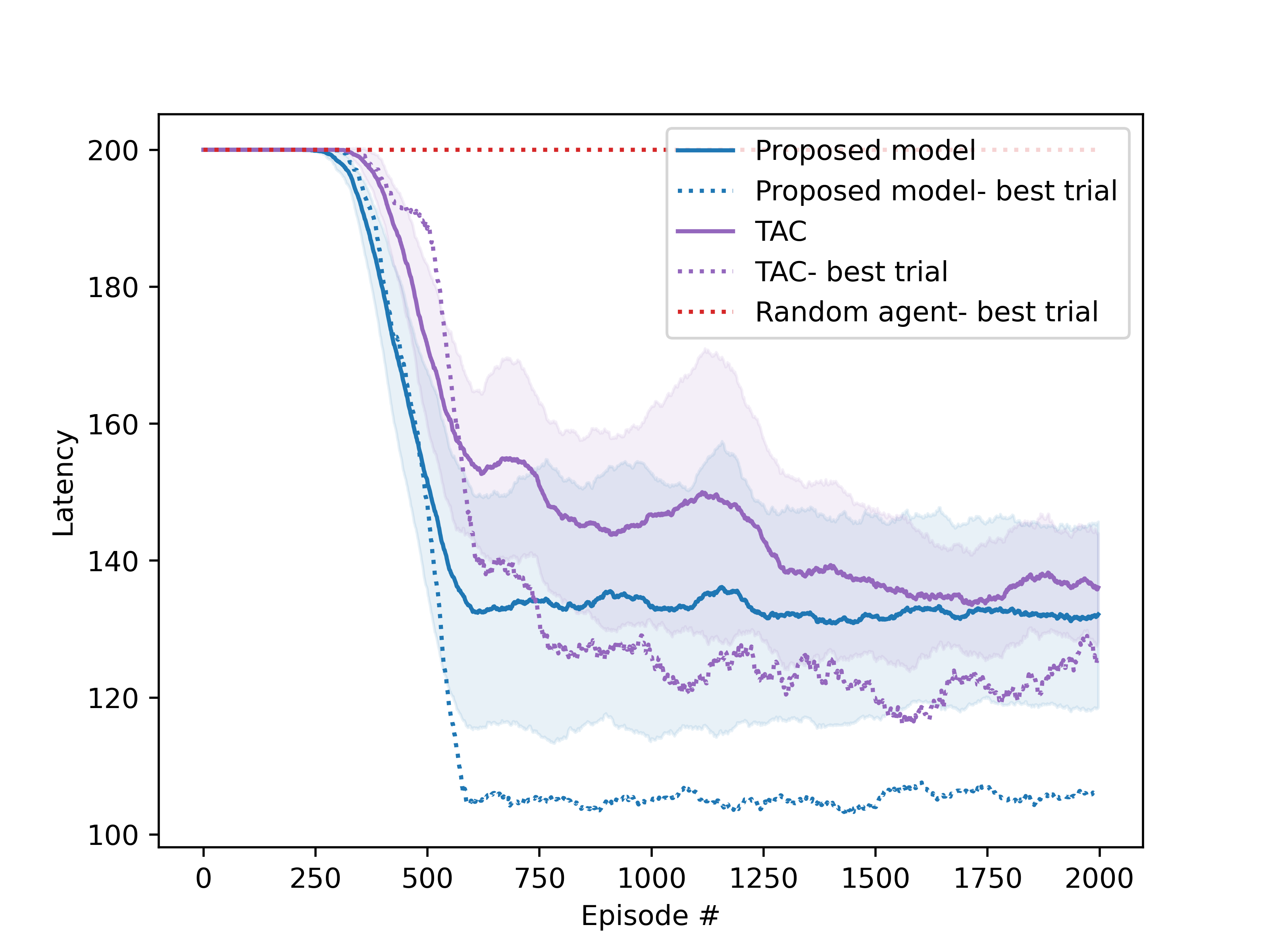}  
  \caption{Average latency results on the Mountain car environment. Dotted lines correspond to the best out of ten trials and the shaded regions represent the standard deviation.}  
  \label{fig_res_mountain_car}
\end{figure}

The Mountain car environment has two input dimensions and a sparse reward. The PPO algorithm did not obtain good results on this benchmark unless the reward signal is modified to include a continuous reward related to proximity to the goal state. Because of this, the PPO algorithm is not included in the results presented in Figure \ref{fig_res_mountain_car}. Also due to the sparse reward, a random agent is unable to terminate the episode before the time-out of 200 steps. 

The proposed model and the TAC baseline present similar final latency values and learning speed on average. However, a comparison of the best trials (dotted lines) suggests that the proposed network can surpass the tabular algorithm. It is also worth noting that the observation space of the actor-critic TAC in this environment consists of the entire state space, discretized using 20 bins, giving a total of 400 discrete states. Meanwhile, the proposed network employs 100 neurons for clustering the real-valued 2D input from the environment, resulting in 100 discrete states for the subsequent actor-critic component. This reduction in the state-space becomes more significant in the next benchmark environments with higher state dimensions. It can also contribute to the faster learning rate observed when comparing both average and best latency curves.

\subsubsection{Cart-pole} \label{ssec_exp_results_Cart-pole}

\begin{figure}[H]
\centering
  \includegraphics[width= 0.8 \linewidth]{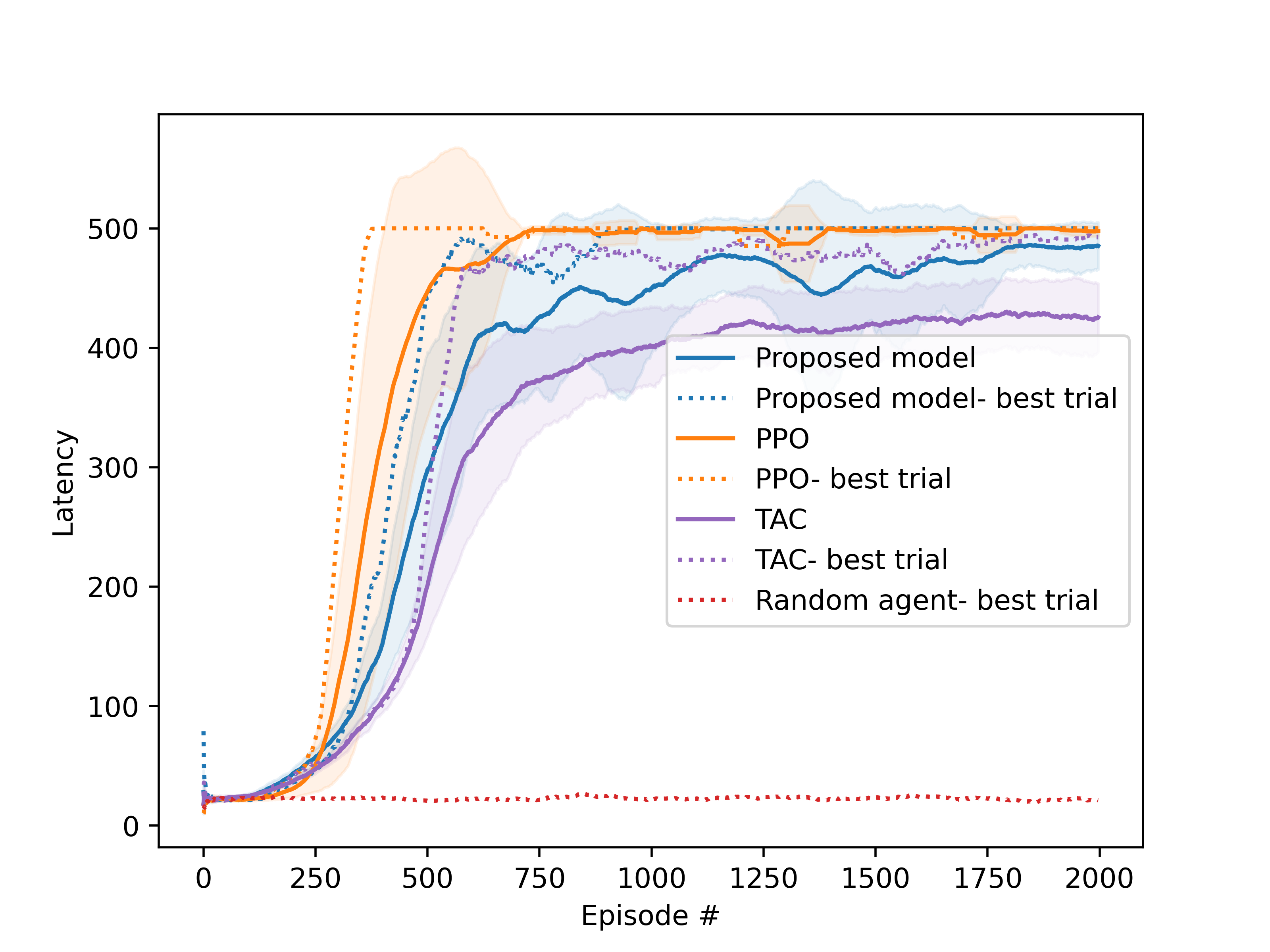}  
  \caption{Average latency results on the Cart-pole environment. Dotted lines correspond to the best out of ten trials and the shaded regions represent the standard deviation.}  
  \label{fig_res_cartpole}
\end{figure}

The proposed model is also compared to baselines on the Cart-pole environment, as illustrated in Figure \ref{fig_res_cartpole}. Differently from the Mountain car problem, the reward is not sparse and is proportional to the time the agent is able to keep the pole in balance. While all three of the evaluated models are able to significantly improve the balancing time after 250 episodes, PPO provides the most stable control over the ten independent runs. The proposed model obtains a comparable performance to PPO on the best run and, as in the previous experiment, is better than TAC in terms of both average and the best runs.

As described in Section \ref{ssec_exp_hyperparams_proposed}, the proposed network uses a single clustering layer with 100 neurons, providing a reduced discrete state space for the actor-critic network. On the other hand, the TAC algorithm uses 10 bins for each input dimension, giving a total of $10^4$ discrete states. While increasing the number of bins can result in improved performance over time, this has the drawback of slower learning speed. Thus, the proposed clustering layer is able to provide enough resolution in the state-space for solving the Cart-pole without requiring a very large number of neurons. 

\subsubsection{Acrobot} \label{ssec_exp_results_acrobot}
The Acrobot environment combines a sparse reward with additional input dimensions, making it the most challenging of the three considered benchmarks. On this task, the proposed network is able to achieve significantly better performance by using two clustering layers. The first layer contains 20 neurons per dimension, or a total of 120 neurons, with 6 active at any given time-step. The second layer reduces this state-space to just 20 neurons in total, a substantial reduction compared to $10^6$ states observed by the TAC algorithm.

\begin{figure}[H]
\centering
  \includegraphics[width= 0.8 \linewidth]{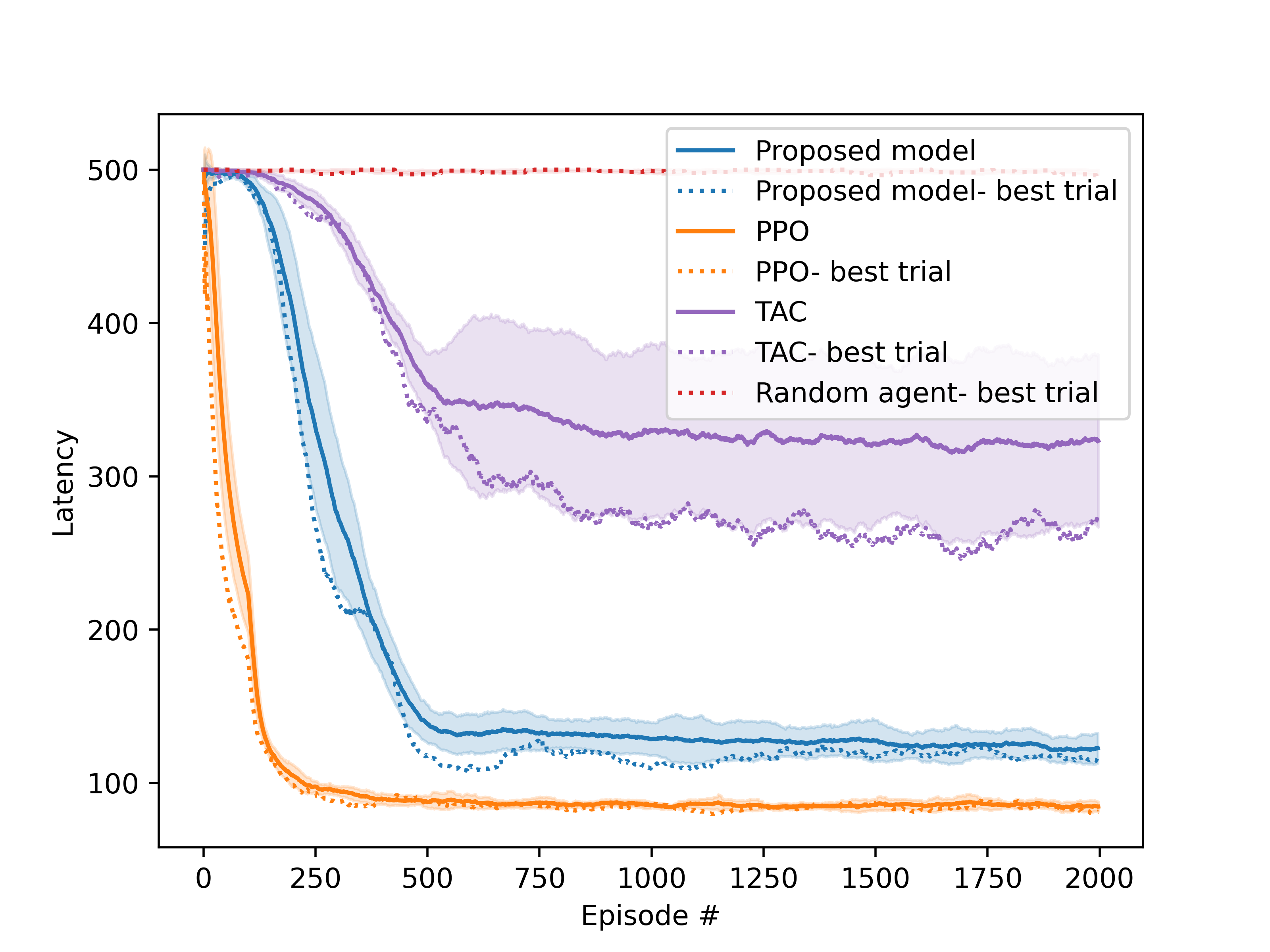}  
  \caption{Average latency results on the Acrobot environment. Dotted lines correspond to the best out of ten trials and the shaded regions represent the standard deviation.}  
  \label{fig_res_acrobot}
\end{figure}

While the PPO algorithm is  significantly better than the proposed one in terms of learning speed and control precision, the proposed model is still able to find a stable control strategy for the Acrobot problem.     

\subsection{Discussion and summary} \label{sssec_discussion}
An obvious advantage of the proposed network over a tabular RL algorithm is that the state space observed by the actor is drastically reduced. Table \ref{tab_comparison_td_lambda} provides a comparison of the proposed model and the Tabular Actor-Critic (TAC) in terms of state space requirements. The difference becomes more significant when the number of input dimensions is increased.  

\begin{table}[htbp]
  \small
  \centering
  \caption{Comparison between the proposed model and the Tabular Actor-Critic (TAC) in terms of state space requirements}
  \label{tab_comparison_td_lambda}
  \begin{tabular}{
    @{}
    l
    c
    S[table-format=1.0e1]
    @{}
  }
    \toprule
    \textbf{Environment} & \textbf{Algorithm} & {\textbf{State space}} \\ 
    \midrule
    Mountain car & Proposed   & 1e2 \\
        & TAC & 4e2   \\
    \midrule
    Cart-pole    & Proposed   & 1e2 \\
        & TAC & 1e4 \\
    \midrule
    Acrobot      & Proposed   & 2e1 \\
        & TAC & 1e6\\
    \bottomrule
  \end{tabular}
\end{table}

We consider the Cart-pole and Acrobot environments for an additional ablation study with the aim to evaluate the impact on the performance of the individual components of the proposed network. In order to provide a more statistically significant set of results, each experiment consists of 30 independent runs using a specific configuration of the algorithm. Average latency and standard deviation are evaluated using an unpaired t-test with 95\% significance. 

\begin{figure}[htbp]
\centering
  \includegraphics[width= 0.8 \linewidth]{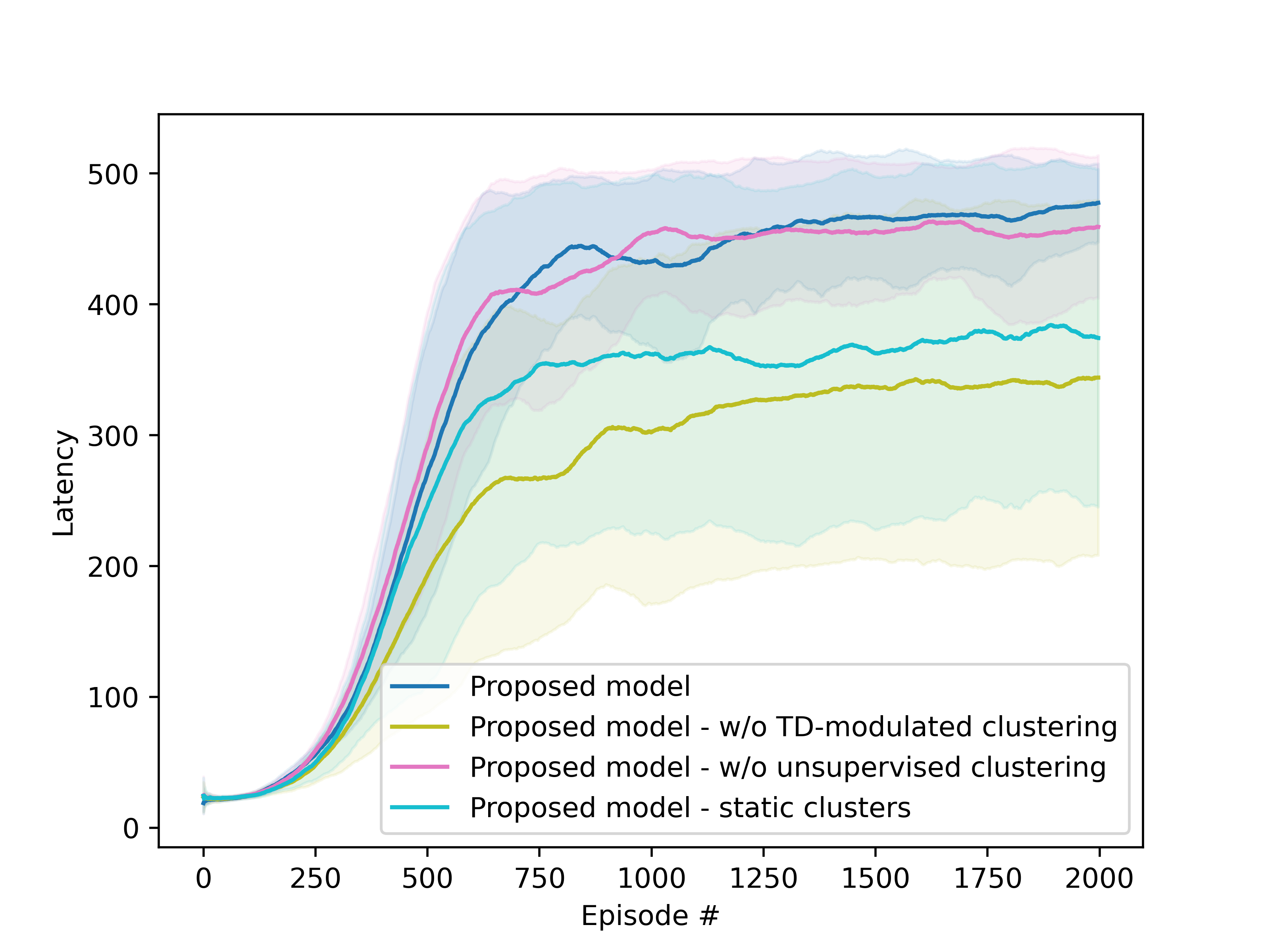}  
  \caption{An ablation study of the proposed model on the cart-pole environment. The shaded regions represent the standard deviation.}  
  \label{fig_cart_pole_ablation}
\end{figure}

The proposed network has two synaptic plasticity models for the clustering layers: passive clustering that drives neurons to evenly represent the observed state-space and TD-error modulation that encourages a denser distribution near states with higher absolute TD-error (see Section \ref{ssec_proposed_SNN_ac_layer} for more details). Each of these components is individually disabled from the proposed model and evaluated on the cart-pole environment. The average latency and standard deviation are calculated from the last 1000 episodes of each run, once a more stable control is reached. A qualitative comparison is presented in Figure \ref{fig_cart_pole_ablation}, while the statistical significance analysis is based on corresponding data from Table \ref{tab_cart_pole_ablation}. 

The comparison of average latencies in Table \ref{tab_cart_pole_ablation} suggests that TD-error modulation plays a significant role in the performance of the network. Meanwhile, there is no statistically significant difference between a configuration with and without passive clustering enabled on the cart-pole environment. Interestingly, both plasticity rules are found to be significant in the next ablation study on the acrobot environment.

\begin{table}[htbp]
  \small
  \centering
  \caption{Average and standard deviation of the latency curve for the ablation study on the cart-pole environment. The shaded regions represent the standard deviation.}
  \label{tab_cart_pole_ablation}
  \begin{tabular}{
    @{}
    l
    c
    c
    @{}
  } 
    \toprule
    \textbf{Configuration} & \textbf{Average latency (std)} & {\textbf{Significant change?}} \\ 
    \midrule
    Proposed & 460 (52) & \\
    W/o TD-modulated clustering & 332 (134) & \checkmark \\
    W/o unsupervised clustering & 456 (56) &  \\
    Static clusters & 367 (132) & \checkmark \\
    \bottomrule
  \end{tabular}
\end{table}

\begin{figure}[htbp]
\centering
\begin{subfigure}{.7\textwidth}
  \centering
  \includegraphics[width=\textwidth]{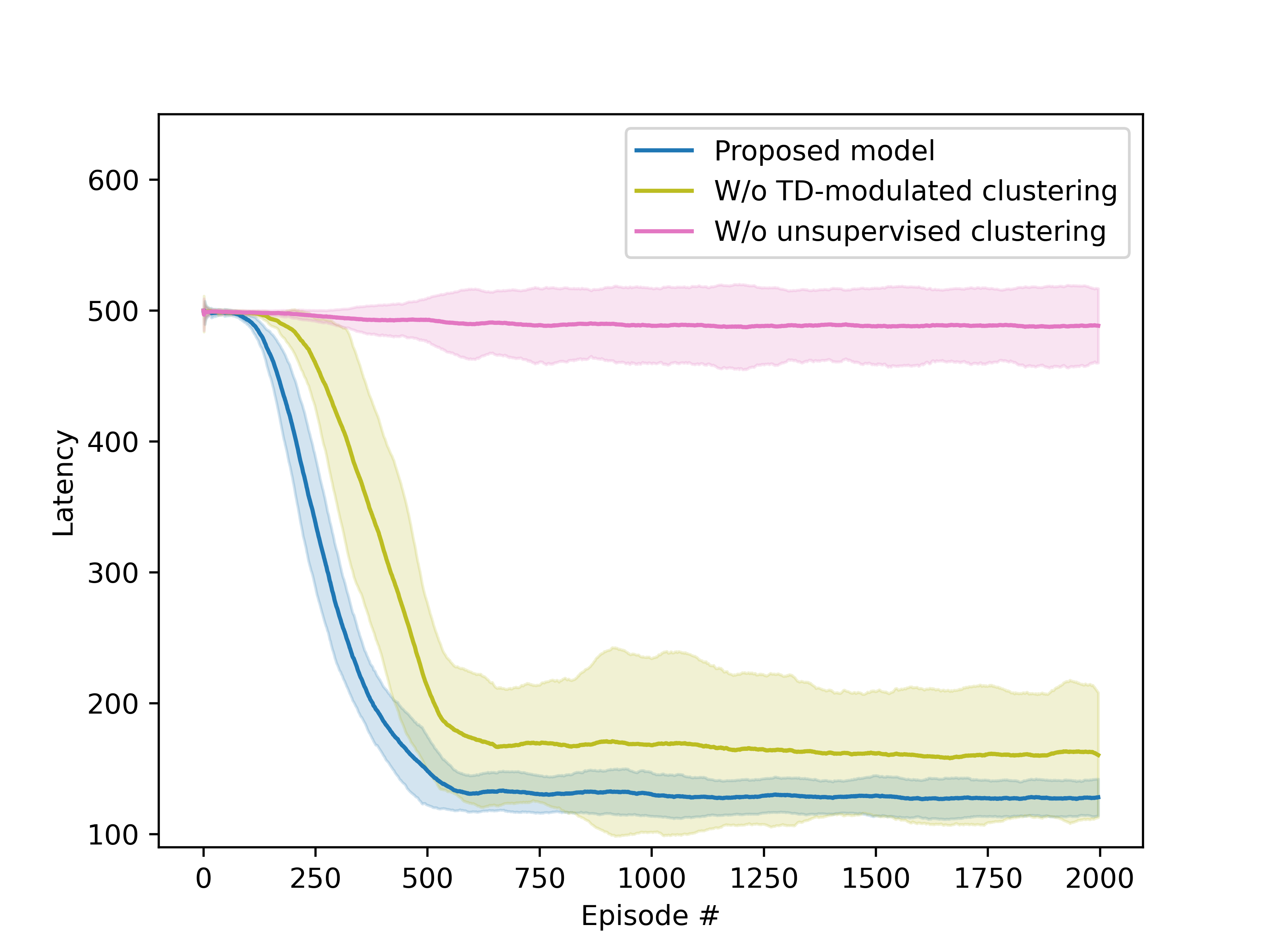}
  \caption{}
  \label{fig_results_acrobot_ablation_1}
\end{subfigure}~ 
\newline
\begin{subfigure}{.7\textwidth}
  \centering
  \includegraphics[width=\textwidth]{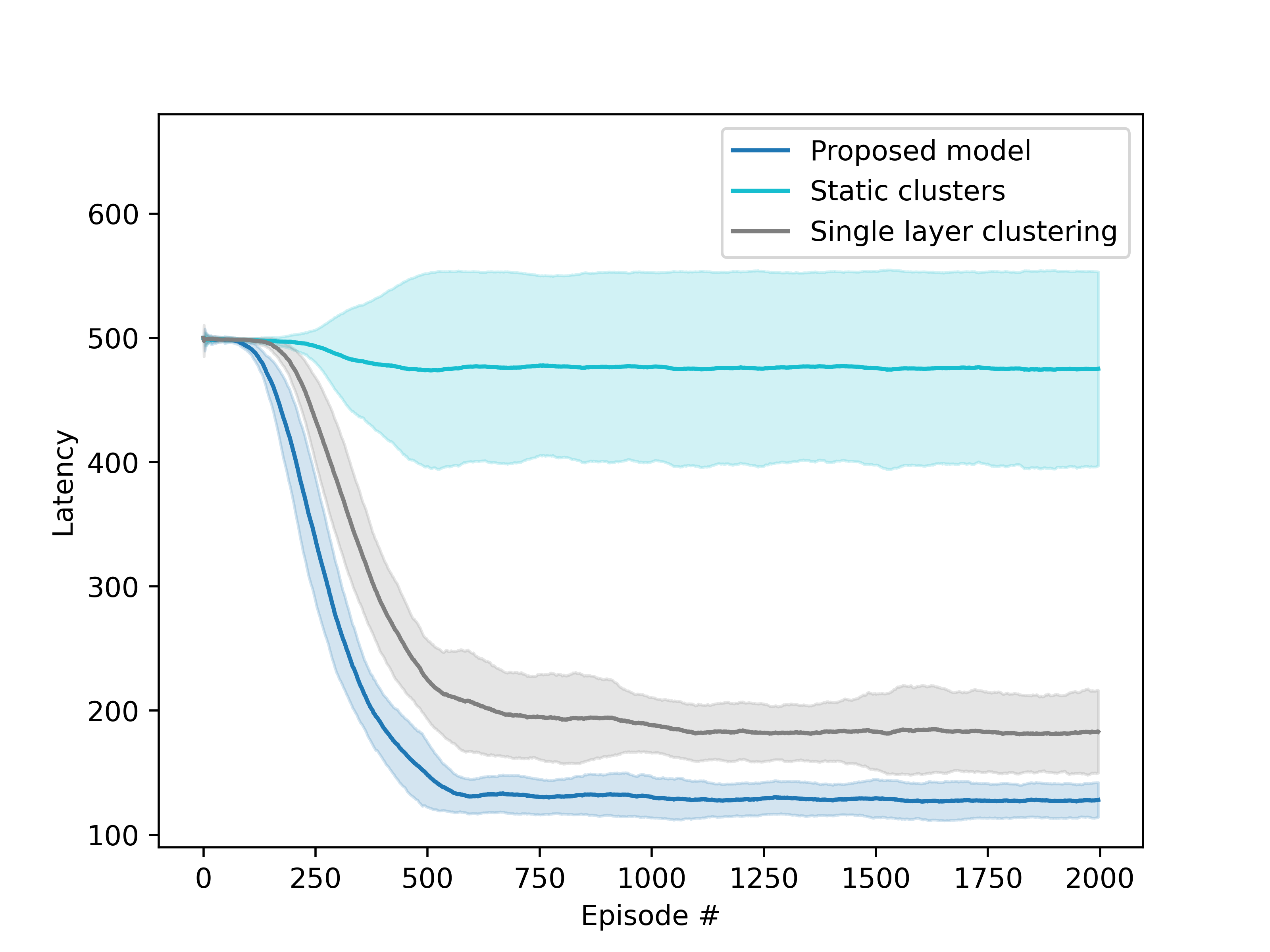}
  \caption{}
  \label{fig_results_acrobot_ablation_2}
\end{subfigure}~
\newline
\caption{An ablation study of the proposed model on the acrobot environment. The shaded regions represent the standard deviation.}
\label{fig_acrobot_ablation}
\end{figure}

A similar ablation study on the acrobot environment is presented in Figure \ref{fig_acrobot_ablation} and Table \ref{tab_acrobot_ablation}. However, since the proposed network for this environment has a second clustering layer, another configuration is optimized with a single clustering layer in order to evaluate the impact of this feature. Simply removing a layer from the proposed configuration would not result in a fair comparison, since the other hyperparameters also have to be adjusted to better fit the new configuration. Thus, an additional optimization run with 500 trials is performed, considering the same range of hyperparameters presented in Table \ref{tab_hyper_proposed}.    

Overall, each of the considered changes to the proposed architecture results in a statistically significant drop in performance, when compared to the initial configuration. In contrast to the previous experiment with the cart-pole environment, the unsupervised clustering plays a more significant role than the TD-modulated one. 

\begin{table}[htbp]
  \small
  \centering
  \caption{Average and standard deviation of the latency curve for the ablation study on the acrobot environment}
  \label{tab_acrobot_ablation}
  \begin{tabular}{
    @{}
    l
    c
    c
    @{}
  }
    \toprule
    \textbf{Configuration} & \textbf{Average latency (std)} & {\textbf{Significant change?}} \\ 
    \midrule
    Proposed & 128 (14) & \\
    Static clusters & 475 (78) & \checkmark \\
    Single clustering layer & 183 (29) & \checkmark \\
    W/o TD-modulated clustering & 163 (54) & \checkmark \\
    W/o unsupervised clustering & 488 (29) & \checkmark \\
    \bottomrule
  \end{tabular}
\end{table}

Finally, the proposed neuromorphic approach is fundamentally different from traditional deep reinforcement learning (DRL) models in term of both memory usage and the amount of synaptic updates that are applied to the network during learning. Firstly, while PPO and similar DRL algorithms use a large memory buffer for storing previous transitions in the environment, the current approach relies on synaptic traces. Secondly, the backpropagation algorithm requires global information about the network, while in the proposed model only local computations are applied for synaptic plasticity. Thirdly, during training the synapses of the actor-critic network used by PPO are modified several times over a batch of data retrieved from the memory buffer. On the other hand, the synaptic updates on the proposed model are performed once at each time-step by using a broadcasted TD-error signal and eligibility traces.

\section{Conclusion and final remarks}
\label{sec_conclusion}
This work provides a novel neuromorphic architecture aimed at solving reinforcement learning problems with real-valued observations. The proposed network contains clustering layers, based on earlier work by \cite{afshar2020event} and \cite{BethiEtAl2022}, with an introduction of TD-error modulation and eligibility traces. The impact of the main components introduced in this architecture is evaluated in an ablation study and shown to have a significant impact on performance.

The network's effectiveness is assessed against a tabular actor-critic algorithm with eligibility traces, as well as a state-of-the-art deep learning model, Proximal Policy Optimization (PPO). The proposed model outperforms the tabular algorithm in terms of learning speed and accuracy consistently, demonstrating its capability to discover stable control policies for three control problems involving 2, 4 and 6 input dimensions. 

While it does not surpass the PPO-based controller in terms of optimal performance, our network offers an appealing trade-off in terms of memory and hardware  implementation requirements. This is because the proposed model does not require an external memory buffer and the synaptic plasticity occurs online,  driven solely by local learning rules and a broadcasted TD-error signal. These aspects provide a more biologically plausible learning model, which in future may lead to more efficient algorithms in terms of computational and memory requirements. 

\subsection{Future work}
\label{ssec_future_work}
Future research can focus on the following directions:

\begin{itemize}
    \item Automatic parameter and architecture tuning -- the experiments suggest that the model's performance is sensitive to the choice of several parameters. Future work could discover how to best apply evolutionary algorithms to the problem of finding architecture and parameter set for a given problem.
    \item Scalability to more complex problems -- the proposed model was effective in solving control problems with 2, 4, and 6 input dimensions. Future work could investigate the model's scalability to problems with even larger input dimensions. A hierarchic organization in the clustering layers is one possible strategy for learning from an increased number of dimensions.
    \item Hardware implementation -- the study highlighted the potential hardware efficiency of the proposed model. Future work could focus on the actual implementation of this architecture in hardware, and investigate its performance in real-world applications.
    \item Population coding -- the current method does not rely on population coding, which can increase the number of parameters within the network and lead to slower learning speeds. However, more robust learning and better scalability are potential benefits of this encoding method \parencite{tang2021deep}. Future work could explore a hybrid approach that combines the current method with elements of population coding, to potentially improve performance without significantly increasing the number of parameters.   
\end{itemize}

\section*{Acknowledgment}
The  authors  would  like  to  thank  FACEPE, CNPq  and CAPES (Brazilian Research Agencies) for their financial support.

% \appendix
% \section{Supplementary material} \label{sec_suplementary}

\printbibliography

\end{document}